\documentclass[letterpaper, 10 pt, conference]{ieeeconf}  

\IEEEoverridecommandlockouts   

\usepackage{graphics} 
\usepackage{epsfig} 
\usepackage{amsmath} 
\usepackage{amssymb}  
\usepackage{mathtools}
\usepackage{booktabs}
\usepackage{multirow}
\usepackage{hyperref}
\graphicspath{{images}}
\usepackage{nicefrac, xfrac}
\usepackage{todonotes}
\usepackage{fancyhdr}

\title{\LARGE \bf
Enhancing LLM-Based Human-Robot Interaction with Nuances\\for Diversity Awareness}

\author{Lucrezia Grassi,
Carmine Tommaso Recchiuto,
Antonio Sgorbissa
\thanks{All authors are with RICE lab at DIBRIS, University of Genoa, Via all'Opera Pia 13, 16145 Genoa, Italy.}
\thanks{Corresponding author's email: \texttt{\href{mailto:lucrezia.grassi@edu.unige.it}{lucrezia.grassi@edu.unige.it}}}%
}

\fancypagestyle{plain}{
    \fancyhf{}

    \fancyfoot[C]{\footnotesize
    \textit{© 2024 IEEE. Personal use of this material is permitted. Permission from IEEE must be obtained for all other uses, in any current or future media, including reprinting/republishing this material for advertising or promotional purposes, creating new collective works, for resale or redistribution to servers or lists, or reuse of any copyrighted component of this work in other works. This article has been accepted for publication in IEEE ROMAN 2024. The final published version is available at \textit{[DOI will be inserted here once available]}.
    }
    }
}

\begin{document}

\pagestyle{plain}

\maketitle
\thispagestyle{plain} 

\begin{abstract}
This paper presents a system for diversity-aware autonomous conversation leveraging the capabilities of large language models (LLMs). The system adapts to diverse populations and individuals, considering factors like background, personality, age, gender, and culture. The conversation flow is guided by the structure of the system's pre-established knowledge base, while LLMs are tasked with various functions, including generating diversity-aware sentences. Achieving diversity-awareness involves providing carefully crafted prompts to the models, incorporating comprehensive information about users, conversation history, contextual details, and specific guidelines. To assess the system's performance, we conducted both controlled and real-world experiments, measuring a wide range of performance indicators.
\end{abstract}

\section{Introduction}
With the increased adoption of Socially Assistive Robots (SARs), there is a growing need for these robots to not only exhibit cultural competence but also embrace a broader perspective of diversity. Diversity, in this context, encompasses unique individual qualities, including culture, personality, beliefs, and more. Customizing systems to individual needs is crucial to prevent the generation of discomfort and ensure well-being, particularly for vulnerable users.

Researchers in the field of Artificial Intelligence (AI) and robotics have made efforts to tailor robots to individual needs \cite{martin2020}. One approach involves employing machine learning (ML) to gather and analyze extensive data periodically, aiming to gain insights into the individual and their environment \cite{parisi2019}. However, relying solely on data collection and processing is unrealistic for capturing all individual nuances. Furthermore, these algorithms may produce biased outputs due to programmers' unconscious biases \cite{buolamwini2018, bolukbasi2016}. The emerging field of “fair'' ML addresses these concerns and shares similarities with diversity awareness \cite{mehrabi2021}. Fairness focuses on perception and decision-making, while diversity-aware robots prioritize real-time interaction by integrating diversity into all aspects \cite{recchiuto2022}.

In recent years, there has been a growing interest in leveraging LLMs to enhance the interaction capabilities of robots. For instance, Microsoft engineers employed ChatGPT for robotics applications, emphasizing prompt engineering and dialogue strategies \cite{vemprala2023chatgpt}. However, LINE Corporation\footnote{\url{https://linecorp.com/en/}} researchers highlighted the limitations of purely LLM-based systems, which can produce irrelevant responses \cite{takato2023}. They developed a dialogue system that breaks tasks into smaller sub-tasks and uses LLMs more effectively.

The integration of GPT-3 with Aldebaran Pepper and NAO robots has facilitated open verbal dialogues \cite{Billing2023}. Researchers are investigating the use of generative language models like GPT-3 to enhance interactive learning with educational social robots acting as tutors and companions \cite{Sonderegger2022}. However, these models must be guided by carefully crafted prompts. These prompts direct LLMs, facilitating adherence to guidelines, task automation, and regulation of specific content aspects. Yet, prompts are often created intuitively and unsystematically by non-AI experts, resulting in unintended and undesirable outputs \cite{zamfirescu2023johnny}. Consequently, the influence of prompts on LLM performance has led to the emergence of a specialized field known as “prompt engineering'' \cite{gao2023, liu2023prompt}. 

In light of these considerations, the research question is the following: \textit{can a system be designed to adapt the conversation to individual characteristics, based on the notion of diversity-awareness, enhancing the user experience and mitigating feelings of discomfort, while harnessing the capabilities of large language models?}

\begin{figure}
	\centering
	\includegraphics[width=\linewidth]{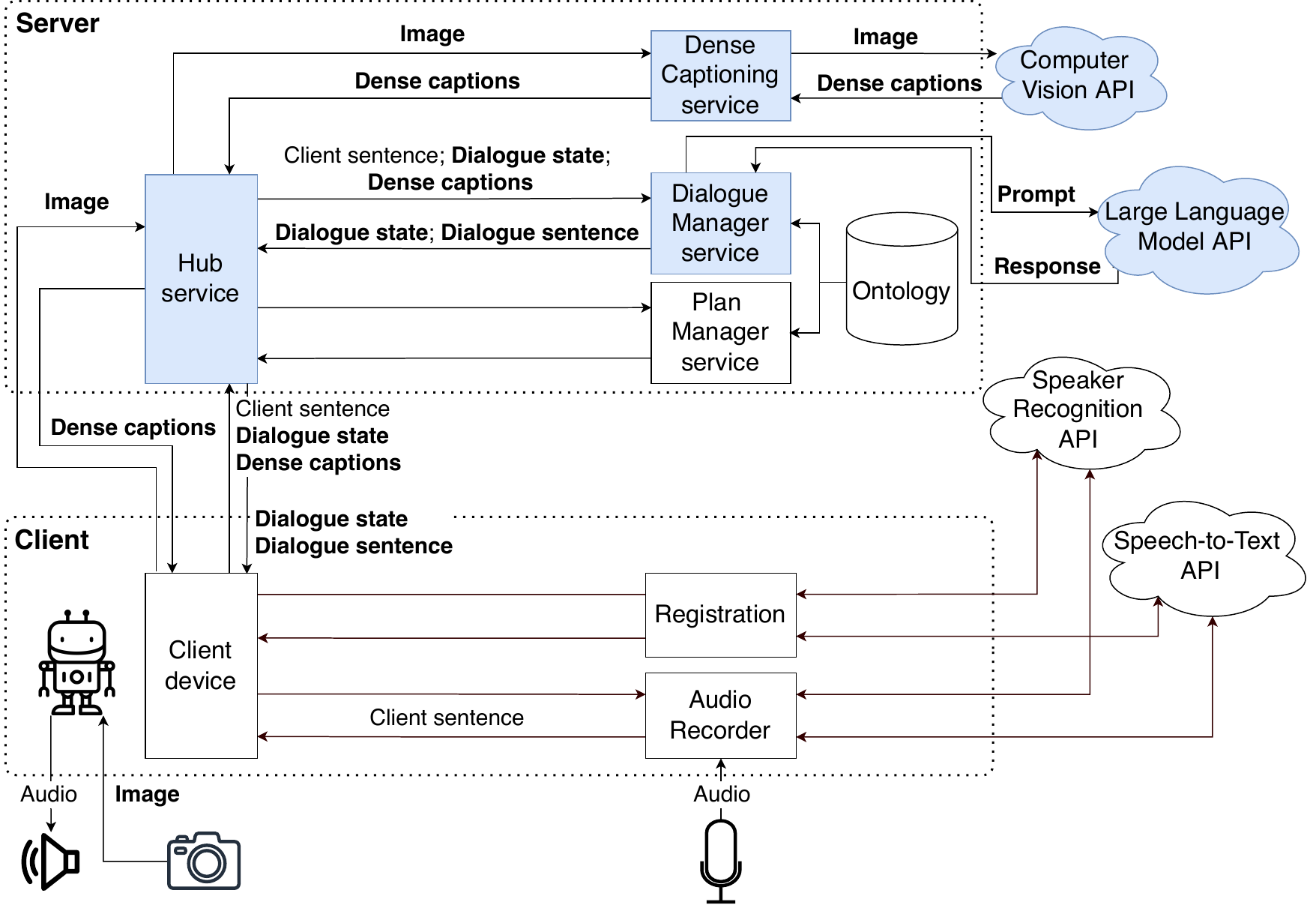}
	\caption{CAIR system architecture integrating LLMs and dense captioning. Modified or added blocks are colored in blue, while modified or added messages are highlighted in bold.}
	\label{fig:architecture_llm_vision}
 \vspace{-4mm}
\end{figure}

This work delves into the notion of using large language models to adapt the conversation to the individuals, while the assessment of the user experience will be addressed in future investigations. To this aim, the study presents the architecture of a cloud system for diversity-aware autonomous interaction, relying on an ontology for knowledge representation and dialogue management. The primary contribution of this research lies in the definition of a methodology to leverage the capabilities of large language models for tailoring conversations, focusing on what we term “dialogue nuances.''

The work is structured as follows: Section \ref{sec:system-architecture} provides an overview of the system's architecture. 
Section \ref{sec:experiments} delves into the experiments conducted to assess the performance of the system and discusses its initial deployment in real-world settings. 
Eventually, Section \ref{sec:conclusions} draws the conclusions.

\section{System architecture}
\label{sec:system-architecture}
The system presented in this paper is a modification of the CAIR (Cloud AI and Robotics) system described in \cite{grassi2023iros}. CAIR is a cloud-based system for autonomous interaction built upon an OWL2 ontology for rich, knowledge-grounded conversations \cite{recchiuto2020b}. The ontology is designed to consider cultural differences between users in a non-stereotyped manner. It stores conversation topics and pre-defined sentences, enabling dynamic composition at runtime and facilitating culturally aware and engaging conversations \cite{recchiuto2020a}.

Figure \ref{fig:architecture_llm_vision} illustrates the modified architecture of the system. Blue blocks and bold text highlight the changes compared to previous works \cite{grassi2023iros, grassi2023jist}. It is important to note that this paper does not discuss system components related to plan management, speaker registration for multi-party interaction, or the details of audio acquisition and speaker recognition, as these aspects have been addressed in previous publications. Instead, this work focuses on the modifications performed to the Dialogue Manager service, responsible for conversation management, to obtain diversity-aware sentences generated by LLMs instead of retrieving them from the ontology. Additionally, to enhance diversity awareness by grounding conversations in visual information, the solution relying on dense captioning for object retrieval and their relationships \cite{grassi2024ICRA} has been integrated into the CAIR system.

Dialogue management and visual information acquisition occur simultaneously through two parallel threads. These threads are initiated by the main client thread, which oversees the robot's speech and movement. A separate thread oversees audio acquisition. Upon receiving the user's sentence from the audio acquisition thread, the main thread triggers the execution of the dialogue thread, responsible for managing requests to the CAIR server. Meanwhile, the vision thread continuously updates visual information. 

The system currently leverages models provided by OpenAI APIs due to their superior performance over open-source alternatives. However, the proposed architecture is flexible and can be easily adapted to integrate any language model.

\subsection{The role of the ontology}
Employing an LLM involves more than simply inputting a user sentence and receiving the model's response. The process is complex, especially when the aim is to retain control over the conversation. In this work, this is achieved by following the structure of the knowledge base, considering the current conversation topic and desired sentence type, and adhering to predefined patterns for each topic. Greater control over the conversation flow ensures a higher respect for individual diversities, adapting to each person's needs and preferences, while keeping the conversation aligned with ontology topics and avoiding unwanted digressions.

\subsection{Dialogue nuances}
\label{sec:dialogue_nuances}
Prompt design is crucial for ensuring diversity-aware content. This necessity led to the inclusion of additional information in the prompt, consistently provided to the model. To this aim, the \textit{dialogue state} \cite{grassi2023jist}, containing information about the state of the conversation in terms of covered topics, user preferences, and conversation patterns, has been extended to encompass what is referred to as “dialogue nuances.'' These nuances incorporate details about the individual and conversation guidelines, enhancing sentence generation and controlling model responses. The existing nuances and their corresponding fields include:

\begin{itemize}
	\item Diversity: nationality, mental condition, physical condition;
	\item Time: time of the day, season, events;
	\item Place: environment, city, nation;
	\item Tone: humorous, kind, dramatic, controversial, aggressive, teasing, alarmist, worried;
	\item Speech act: assertive, commissive, expressive, directive.
\end{itemize}

To detail the implementation, the \textit{dialogue state} comprises two structures related to the \textit{dialogue nuances}. The first structure contains vectors representing the values of each nuance. A nuance value vector $\mathbf{v}_k \in \mathbb{R}^m$ is defined as $\mathbf{v}_k := [v_{k,1}, \ldots, v_{k,m}]^\top$, where $k \in \{d, t, p, n, s\}$ refers to a specific nuance, respectively diversity ($d$), time ($t$), place ($p$), tone ($n$) and speech act ($s$), and where each element $\mathbf{v}_{k, i}$, $i=1, \dots, m$ represents the specific value of the nuance. More in detail, the values for each nuance value vector are: 
\begin{equation*}
	\begin{aligned}
		\mathbf{v}_d =& [\text{$<$nationality$>$}, 	\text{$<$mental\_condition$>$}, \\ &\text{$<$physical\_ condition$>$}]^\top,\\
		\mathbf{v}_t =& [\text{$<$time\_of\_the\_day$>$}, 	\text{$<$season$>$}, \text{$<$events$>$}]^\top,\\
		\mathbf{v}_p =& [\text{$<$environment$>$}, \text{$<$city$>$}, 	\text{$<$nation$>$}]^\top,\\
		\mathbf{v}_n =& [\text{humorous}, \text{kind}, 	\text{dramatic}, \text{controversial}, \text{aggressive},\\
  &\text{teasing}, \text{alarmist}, \text{worried}]^\top,\\
		\mathbf{v}_s =& [\text{assertive}, \text{commissive}, \text{expressive},
        \text{directive}
  ]^\top.
	\end{aligned}
\end{equation*}

For diversity, time, and place nuances, the values are specific to the individual interacting with the system, while the values of the tone and speech act nuances remain constant across different individuals and correspond to the fields of the nuances. Currently, the nuance values are defined based on the robot's deployment location and are static. Examples of vectors with specific values for the first three nuances are:
\begin{equation*}
	\begin{aligned}
		\mathbf{v}_d &= [\text{Italian}, \text{good mental health}, 	\text{good physical health}]^\top,\\
		\mathbf{v}_t &= [\text{evening}, \text{winter}, \text{almost 	Easter}]^\top,\\
		\mathbf{v}_p &= [\text{house}, \text{Genoa}, \text{Italy}]^\top.
	\end{aligned}
\end{equation*}

The second structure in the \textit{dialogue state} related to the \textit{dialogue nuances} consists of vectors representing flags for each nuance. These nuance flag vectors $\mathbf{f}_k \in \mathbb{N}^{n}$ are defined as $\mathbf{f}_k := [f_{k,1}, \ldots, f_{k,n}]^\top$, where each element $f_{k,i} \in \{0, 1\}$, $i=1, \dots, n$ with $n=m+1$. Additionally, it holds that $\sum_{i=1}^{n} f_{k,i} = 1,$ indicating that only one element is set to one while the rest are zeros. These flags determine whether the system should utilize the corresponding nuance value during response generation. Notably, each flag vector includes an extra element at the end that does not correspond to any nuance value. If all flags associated with a nuance's values are zero while the last flag is one, it indicates that the model can decide whether to incorporate information from that nuance. However, for the tone nuance, when the last unpaired flag is one, the system is directed to maintain a “neutral'' tone, preventing abrupt tone changes. Initially, the flags for all the nuances are set on the client device.

During the interaction, the nuance flags change based on transition matrices defined on the server side for each nuance. This process is clarified by introducing a vector of nuance probabilities $\mathbf{p}_k \in \mathbb{R}^{n}$, where each element $p_{k,i} \in [0, 1]$ denotes the probability of using the corresponding nuance value in the next conversation turn. Thus, $\sum_{i=1}^{n} p_{k,i} = 1.$
The nuance probability vectors are derived from transition matrices $\mathbf{T}_k \in \mathbb{R}^{n \times n}$, where each element $T(j, i)$ indicates the probability of transitioning from nuance value $v_{k,j}$ to $v_{k,i}$. Thus, at time $t+1$, $\mathbf{p}_k (t+1):=\mathbf{T}_k  \mathbf{f}_{k}(t)$.
These probability vectors determine the values of the flag vector $\mathbf{f}_{k} (t+1)$. Each element $p_{k,i}$ in $\mathbf{p}_k$ represents the probability of setting $f_{k,i}$ to one, while all other flags are set to zero. One flag $f_{k,i}$ is then randomly selected to be set to one based on the distribution defined by $\mathbf{p}_k$. This selection process can be expressed as: $f_{k,i} (t+1) \sim \mathbf{p}_k(t+1)$.
Understanding how transition matrix numbers influence flag evolution and nuance value usage entails considering the steady state within a Markov chain framework. The steady state represents the long-term probabilities of the system being in each state. In the case of a nuance, determining its steady-state distribution involves computing the normalized eigenvector $\hat{\mathbf{e}}_{k}$ corresponding to the eigenvalue 1 for the transition matrix $\mathbf{T}_k$. 
In the steady state, $\hat{\mathbf{e}}_{k}$ values represent the probabilities of each nuance value being used in prompt generation. 

The flag vectors are updated for all nuances whenever the client sends a request to the server. This update happens twice for each conversation turn, unless the system detects either an “aggressive'' or “humorous'' tone in the user's sentence. In such instances, the tone nuance flag vector is directly updated, setting the flag corresponding to the detected tone to one, bypassing the aforementioned procedure.

Note that the values and flags linked to the nuances' fields can be tailored according to individual traits and preferred interaction styles. Additionally, new nuances can be introduced into the \textit{dialogue state} as needed.

\subsection{Requests performed to the large language model}
\label{sec:requests-to-the-llm}
\begin{figure}
	\centering
	\includegraphics[width=\linewidth]{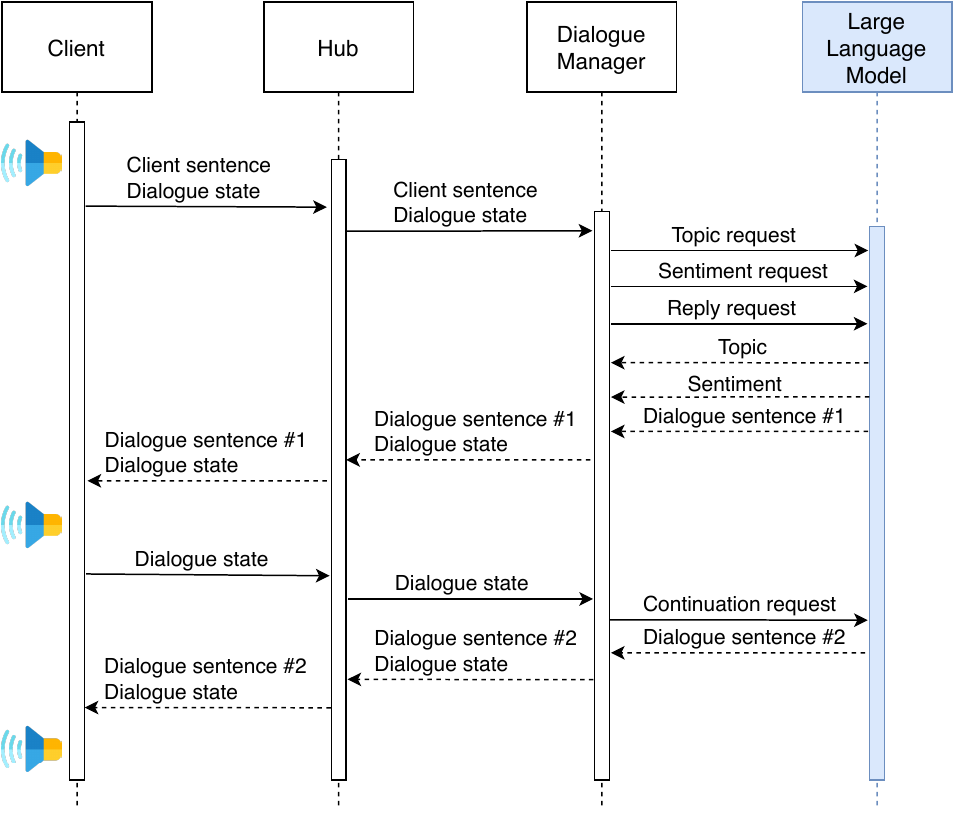}
	\caption{Sequence diagram depicting the requests performed from the Dialogue Manager to the LLM.}
	\label{fig:sequence-diagram-llm}
 \vspace{-5mm}
\end{figure}

To integrate LLMs into the Dialogue Manager, various adjustments were necessary, as illustrated in Figure \ref{fig:sequence-diagram-llm}. The sequence diagram begins with the client responding to the user's input, with speaker icons indicating the client's contributions to the conversation. The initial speaker instance represents the client uttering a brief \textit{filler sentence}, signaling comprehension and initiating the process to obtain the model's response. Many filler sentences, like “Let me think...'' and “I'm still reflecting...,'' applicable to any user input, are pre-generated offline using ChatGPT based on GPT-3.5 to minimize potential repetitions. OpenAI recently introduced a feature enabling developers to receive a continuous stream of model responses. However, this feature is primarily beneficial for longer responses, whereas our prompts explicitly request concise replies to prevent annoyance.

While uttering the first sentence, the client sends the initial request to the Hub, including the \textit{client sentence} and the \textit{dialogue state}. The Hub forwards this request to the Dialogue Manager, which updates the nuance flag vectors and initiates three simultaneous requests to the LLM.

The first request, as shown in Figure \ref{fig:sequence-diagram-llm}, is the “topic request.'' In contrast to the previous system version, which relied on keyword matching, the presented system prompts the language model to identify the topic of the client's sentence from a predefined list of topics sourced from the ontology. The Dialogue Manager's second request to the LLM, denoted as “sentiment request'' in Figure \ref{fig:sequence-diagram-llm}, is designed to determine the sentiment of the client's sentence following a “yes/no question'' posed by the system. LLMs are tasked with identifying a positive or negative sentiment within the user's response whenever possible. The sentiment information is then used to adjust the user's preferences regarding specific topics in the ontology. Concurrently, the Dialogue Manager sends a third request to the LLM, indicated as the “reply request'' in the figure, to generate a response sentence for the client, forming the first part of the \textit{dialogue sentence}.

Once all three requests are fulfilled, the Dialogue Manager updates the \textit{dialogue state} and returns the initial part of the \textit{dialogue sentence}, which the client immediately utters. While the client speaks the first part of the \textit{dialogue sentence} (second speaker icon in Figure \ref{fig:sequence-diagram-llm}), it sends a second request to the Hub to obtain the second part of the \textit{dialogue sentence}, meant to continue the conversation on a specified topic.

To generate the second part of the \textit{dialogue sentence}, the Dialogue Manager updates the flag vectors for all nuances and makes the fourth and final request to the LLM, titled the “continuation request''. Upon receiving the response, the client vocalizes it (third speaker icon in Figure \ref{fig:sequence-diagram-llm}), concluding its turn and transitioning to listening to the user's response.

\subsection{Prompt design}
\label{sec:prompt-design}
To understand how \textit{dialogue nuances} are used to influence GPT to produce diversity-aware responses, note that the OpenAI API interface utilizes three prompt fields: “system,'' “user,'' and “assistant,'' respectively defining the system's philosophy, user input, and assistant content. The system field content plays a pivotal role in providing guidelines to the model and information that the generated content should take into account. In this work, various prompt engineering methods have been used, including zero-shot, one-shot, few-shot learning, and the chain of thought (CoT) approach. While some parts of the system field provide instructions without examples, others include one or a few examples. The CoT method forms the foundation of the design, with the entire system field providing a sequence of instructions for the model to follow in generating its response.

\subsubsection{Topic request}
\label{sec:topic request}
In the \textit{topic request}, the system field provides instructions for the model to generate the desired response. We chose GPT-3.5 Turbo for this task due to its suitability for straightforward tasks and cost-effectiveness compared to GPT-4. Specifically, the model is directed to grasp the user's desired discussion topic by selecting from a list provided in the user field accompanying the sentence. This list, sourced from the ontology, includes both broader and more specific topics related to the ongoing conversation, preventing the model from being overwhelmed with numerous choices and potential confusion. The topic returned by GPT is stored in the \textit{dialogue state} for use during the \textit{continuation request}, as described in Section \ref{sec:continuation-request}.

\subsubsection{Sentiment request}
Similar to the \textit{topic request}, this task is also assigned to GPT-3.5 Turbo. Its goal is to evaluate the sentiment expressed in the user's sentence, requiring the model to categorize it as positive, negative, or neutral. This request is performed following a “yes/no question'' sentence type, allowing the system to assess the user's preference regarding the conversation topic and their inclination to further discuss it, thus enabling personalized interaction.

\subsubsection{Reply request}
\begin{figure}[t]
	\centering
	\includegraphics[width=\linewidth]{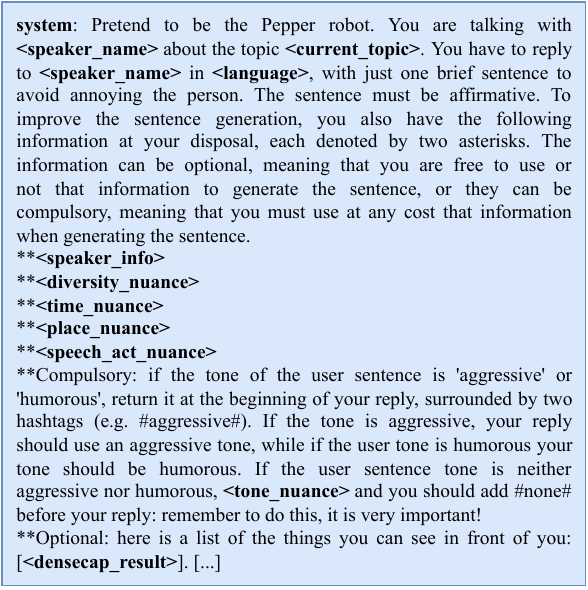}
	\caption{System field of the \textit{reply request}. The content in brackets is replaced by the system with the actual values.}
	\label{fig:reply_system_nuances}
 \vspace{-4mm}
\end{figure}

The purpose of the \textit{reply request} is to obtain the initial part of the \textit{dialogue sentence}, which forms the response to the user's input. This request is more complex than the \textit{topic request} and the \textit{sentiment request}, as its output is influenced by multiple pieces of information. Therefore, the GPT-4 Turbo model has been chosen for this specific task. Figure \ref{fig:reply_system_nuances} presents a simplified version of the system field used for the \textit{reply request}, highlighting all the information subject to variation across conversation turns.

Diversity-awareness in the model's responses is determined by the information listed in the system field, indicated by two asterisks, which include the \textit{dialogue nuances}. Each piece of information is classified as compulsory or optional based on the flags' values, outlined in Section \ref{sec:dialogue_nuances}. If at least one flag of a specific nuance value is set to one, the corresponding information is marked as compulsory in the system field, indicating that the model must use it when generating the response. Conversely, if all flags associated with the values of a nuance are zero, the model is free to decide whether to use any information from that nuance.

It is crucial to emphasize that the inclusion of information related to the tone to employ when generating a response is mandatory. This choice ensures that the model consistently aligns its responses with the tone set by the user. As already explained in Section \ref{sec:dialogue_nuances}, if the user's tone is humorous or aggressive, the tone nuance flags are promptly updated, directing the model to respond accordingly. In cases where the user's tone does not fall into these categories, the model follows the natural evolution of the tone nuance flags.

The last piece of information provided to the model concerns vision information, which falls outside the scope of this work and therefore is not discussed further.

\subsubsection{Continuation request}
\label{sec:continuation-request}
The \textit{continuation request} is the request performed to the model to obtain the second part of the \textit{dialogue sentence}. Similar to the \textit{reply request}, it utilizes \textit{dialogue nuances} in its system field. Due to its complexity, the \textit{continuation request} is directed to the GPT-4 Turbo model to enhance content quality.

In contrast to the \textit{reply request}, where the model is required to simply reply to the previous speaker's sentence, here the model is instructed to use a specific type of sentence regarding the upcoming conversation topic. The conversation topic may change either due to GPT identifying a different topic in the user's sentence or as dictated by the structure of the knowledge base. If transitioning to a new topic, the sentence can be a yes/no question or a positive statement, depending on the user's previously expressed preferences. If the topic remains unchanged, two scenarios arise: further exploration elicits an open-ended question or goal proposal, while a thoroughly examined topic may lead to an exhortative sentence encouraging the user to choose a new subject. Notably, for yes/no or open questions, the “directive'' speech act is consistently employed. Conversely, other sentence types align with the flag values to determine the speech act.

OpenAI's prompt structure enables the integration of multiple assistant and user fields, enriching response generation with contextual information that acts as a memory. To achieve context-awareness, the \textit{dialogue sentence} and the user's input are appended to the prompts of reply and continuation requests at each conversation turn. These prompts are then sent to the model to generate the subsequent \textit{dialogue sentence}. Currently, the prompt retains the last five turns in addition to the system field, which changes with each request as the variables are updated.

\section{Experiments and results}
\label{sec:experiments}
\subsection{Diversity-aware robotics: performance tests}
\label{sec:results_diversity_performance}
This section delves into the analysis of data collected from a controlled experiment conducted 
with the NAO robot. The experiment logged extensive data to examine system performance. Moreover, we offer insights into the “diversity cost,'' a metric we introduced to evaluate the expense of integrating diversity awareness into the system in terms of prompt tokens, i.e. the basic units for measuring the length of text processed by the models.

Throughout this experiment, a total of 300 sentences were pre-scripted to be spoken by the experimenter, aiming to elicit responses from the system. These sentences were generated using ChatGPT. For the first 100 sentences, ChatGPT was directed to produce sentences with a humorous tone. For the following 100 sentences, an aggressive tone was required, and for the last 100 sentences, a neutral tone was instructed. Despite being pre-written, these sentences were used within actual conversations with the robot, following its natural flow. Therefore, the impact of the sentences varied depending on the specific phase of the conversation in which they were employed, such as during a yes/no question or an invitation from the robot to delve further into a topic.

\subsubsection{Topic recognition} 
The use of pre-generated sentences pronounced by the experimenter, diverging from natural conversation flow, led to topic recognition 262 times out of 300 sentences (87.33\%), with 242 instances of “topic jumps'' (80.67\%), indicating changes in conversation topics triggered by the user input. Note that the sentences spoken by the user cannot be customized to be an appropriate response to the system's queries since they are pre-written without prior knowledge of the conversation's content. Consequently, these sentences tend to be longer than a typical user's response, frequently prompting the system to transition to a different conversation topic. In total, the \textit{topic request} facilitated the exploration of 48 ontology topics throughout the interaction, resulting in an average \textit{diversity cost} of 329 tokens.

\subsubsection{Sentiment recognition}
The \textit{sentiment request} was executed 165 times out of 300 interaction turns, following a yes/no question posed by the system in the previous turn, contributing to the \textit{diversity cost} with an average of 96 tokens. The recognized sentiments are distributed as follows: neutral ($17.37\%$), positive ($17.96\%$), and negative ($64.67\%$). The prevalence of negative sentiments can be attributed to the explicit instructions given to ChatGPT  when generating the 300 sentences to be spoken by the user. These instructions, which include directives to adopt humorous, aggressive, or neutral tones, indirectly influence the sentiment of the generated responses. More specifically, the high percentage of negative sentiments could stem from the negative tone recognized when an aggressive sentence is used.

\subsubsection{Usage of dialogue nuances}
The data logged enables us to validate whether the update mechanism of the dialogue nuances, based on transition matrices, yields the anticipated steady-state vector. Additionally, we recorded the average number of tokens in the \textit{reply} and \textit{continuation requests}, totaling 794 and 787, respectively, along with the average number of tokens associated with each nuance used in the system field of the prompts. This data, discussed below, enables us to compute the \textit{diversity cost} for each nuance across both requests.

Table \ref{tab:diversity_nuance_values_controlled} presents the distribution percentages of diversity nuance values used in the prompts of both the \textit{reply request} and the \textit{continuation request}. It is important to underscore that the nuances are updated twice at each conversation turn: once before the \textit{reply request} and once before the \textit{continuation request}. Therefore, the percentages are computed over the total number of updates, considering the usage of the nuance values in both the \textit{reply} and \textit{continuation requests}. Note that we expect these percentage to align with the steady-state value, computed as the normalized eigenvector with eigenvalue 1 of the transition matrix. The penultimate column of Table \ref{tab:diversity_nuance_values_controlled} displays the overall usage of the diversity nuance values, computed as the average between the percentages in the first two columns. Remarkably, these values align with the steady-state values in \(\hat{\mathbf{e}}_{d}\), reported in the last column. The average \textit{diversity cost} is 22 tokens in both \textit{reply} and \textit{continuation requests}, representing approximately 2.8\% of the total length of the requests.

\begin{table}[tp]
	\centering
    \caption{Usage of diversity nuance and steady-state distribution.}
	\begin{tabular}{c |p{0.8cm}p{1.5cm}p{1cm}p{0.7cm}}
		\toprule
		Diversity nuance & Reply & Continuation & Overall & \(\hat{\mathbf{e}}_{d}\)\\
		\midrule
		nationality & 8.3\%& 7.0\% & 7.7\% & 0.083\\
		physical condition & 8.0\% & 5.3\% & 6.7\% & 0.083\\
		mental condition & 6.3\% & 7.7\% & 7.0\% & 0.083\\
		free & 77.3\% & 80.0\% & 78.7\% & 0.750\\
		\bottomrule
	\end{tabular}
 \vspace{-4mm}
    \label{tab:diversity_nuance_values_controlled}

\end{table}

The usage percentages of the time and place nuances in the prompts of the \textit{reply} and \textit{continuation requests} are detailed in Tables \ref{tab:time_nuance_values_controlled} and \ref{tab:place_nuance_values_controlled}, respectively. These tables also include their overall usage, which is expected to align closely with the values represented by $\hat{\mathbf{e}}_{t}$ and $\hat{\mathbf{e}}_{p}$, reported in the last column of each table. The average \textit{diversity cost} associated with both the time and place nuances is very similar: 19 tokens for both the reply and continuation requests. This represents roughly 2.4\% of the total length of the requests.

\begin{table}[tp]
	\centering
	\caption{Usage of time nuance and steady-state distribution.}
	\begin{tabular}{c |p{0.8cm}p{1.5cm}p{1cm}p{0.7cm}}
		\toprule
		Time nuance & Reply & Continuation & Overall & \(\hat{\mathbf{e}}_{t}\) \\
		\midrule
		time of the day & 9.3\%& 7.7\% & 8.5\% & 0.082\\
		season & 11.0\% & 9.3\% & 10.2\% & 0.092\\
		events & 8.3\% & 9.7\% & 9.0\% & 0.092\\
		free & 71.3\% & 73.3\% & 72.3\% & 0.735\\
		\bottomrule

	\end{tabular}
   \vspace{-2mm}
    \label{tab:time_nuance_values_controlled}

\end{table}

\begin{table}[tp]
	\centering
	\caption{Usage of place nuance and steady-state distribution.}
	\begin{tabular}{c |p{0.8cm}p{1.5cm}p{1cm}p{0.8cm}}
		\toprule
		Place nuance & Reply & Continuation & Overall & $\hat{\mathbf{e}}_{p}$ \\
		\midrule
		environment & 7.3\%& 9.3\% & 8.3\% & 0.082\\
		city & 6.7\% & 8.3\% & 7.5\% & 0.092\\
		nation & 11.3\% & 10.0\% & 10.7\% & 0.092\\
		free & 74.7\% & 72.3\% & 73.5\% & 0.735\\
		\bottomrule
	\end{tabular}
	\vspace{-4mm}
  \label{tab:place_nuance_values_controlled}
\end{table}

Table \ref{tab:speech_act_nuance_values_controlled}  details the usage of speech act nuance in the prompts of the \textit{reply} and \textit{continuation requests}. Note that, in this scenario, the elements of the transition matrix prevent the usage of the “directive'' speech act, resulting in zero values in both the \textit{reply} vector and the eigenvector $\hat{\mathbf{e}}_{s}$. This restriction arises because directive speech acts, aimed at prompting user action or eliciting information, are unsuitable for the \textit{reply} sentence. Therefore, directive speech acts are exclusively triggered within the \textit{continuation} sentence, typically when the system prompts the user to further elaborate on a given topic. Consequently, only the percentage in the \textit{reply} column aligns with the eigenvector of the transition matrix, diverging from the previous case. The variation in nuance usage between reply and continuation requests reflects in the \textit{diversity cost}. Specifically, the cost of the speech act nuance is higher in the \textit{reply request} compared to the \textit{continuation request}. This is due to a greater percentage of instances where the “free'' value is used, providing the model with complete nuance information. Consequently, the average cost is 11 tokens in \textit{reply requests} (1.4\%) and 8 tokens in \textit{continuation requests} (1\%).

\begin{table}[tp]
	\centering
	\caption{Usage of speech act nuance and steady-state distribution.}
	\begin{tabular}{c |p{0.8cm}p{1.5cm}p{1cm}p{0.7cm}}
		\toprule
		Speech act nuance & Reply & Continuation & Overall & $\hat{\mathbf{e}}_{s}$\\
		\midrule
		assertive & 56.3 \%& 12.7\% & 34.5\% & 0.528\\
		commissive & 11.3\% & 3.0\% & 7.2\% & 0.111\\
		expressive & 9.3\% & 1.7 & 5.5\% & 0.111\\
		directive & 0.0\% & 77.7\% & 38.8\% & 0\\
		free & 23.0\% & 5.0\% & 14.0\% & 0.25\\
		\bottomrule
	\end{tabular}
	 \vspace{-2mm}
  \label{tab:speech_act_nuance_values_controlled}
\end{table}

Table \ref{tab:detected_tones} presents findings on the tones identified in the client sentences. Rows represent the tones specified for ChatGPT to generate the client sentence, while columns show the tones detected by GPT-4 Turbo. From these results it can be seen how the first 100 sentences, generated with a humorous tone, are recognized as humorous 69\% of the times, the successive 100 aggressive sentences have been recognized as aggressive 15\% of the times, while the last 100 neutral sentences have been correctly recognized 99\% of the times. These results, reveal that the detection of neutral sentiment is very accurate, humorous sentences are sometimes interpreted as having neutral tone, while aggressive sentences are rarely detected as aggressive. These discrepancies may stem from limitations in either ChatGPT's generation of humorous content or GPT-4 Turbo's ability to accurately detect humor.

When generating sentences with an aggressive tone, GPT-4 Turbo tends to produce only mildly aggressive content. This limitation arises from OpenAI's policy to prevent the models from generating offensive content, even when explicitly requested. To address this limitation, a potential solution is fine-tuning the models via APIs and providing numerous examples of the content desired. Although this approach has been explored, it revealed that fine-tuning is currently only available for GPT-3.5 Turbo, which is less capable than GPT-4 and GPT-4 Turbo. GPT-3.5 Turbo may struggle to fully comprehend information in the system field, resulting in less appropriate generated content compared to GPT-4. 

An unexplored alternative involves integrating examples of aggressive responses into the prompts of GPT-4 to determine if this prompts the model to generate more aggressive outputs. However, it is essential to consider the potential drawbacks of this approach. Adding more information to already complex prompts may not be cost-effective, and it could adversely impact the accuracy of the model's output. It is worth noting that the current \textit{diversity cost} of the instructions in the system field of the \textit{reply request}, designed to detect user tone, is significant. With 103 tokens, it constitutes approximately 13\% of the prompt length.

\begin{table}[tp]
	\centering
	\caption{Confusion matrix of detected tones.}
	\begin{tabular}{c |ccc}
		\toprule
		\multirow{2}{*}{Client sentence tone} & \multicolumn{3}{c}{Detected tone (\%)} \\
		& Humorous & Aggressive & Neutral \\
		\midrule
		Humorous & 69.0\% & 2.0\% & 29.0\% \\
		Aggressive & 44.0\% & 15.0\% & 41.0\% \\
		Neutral & 1.0\% & 0.0\% & 99.0\% \\
		\bottomrule
	\end{tabular}
 \vspace{-4mm}
	\label{tab:detected_tones}
\end{table}

\begin{figure}[tp]
	\centering
	\includegraphics[width=\columnwidth]{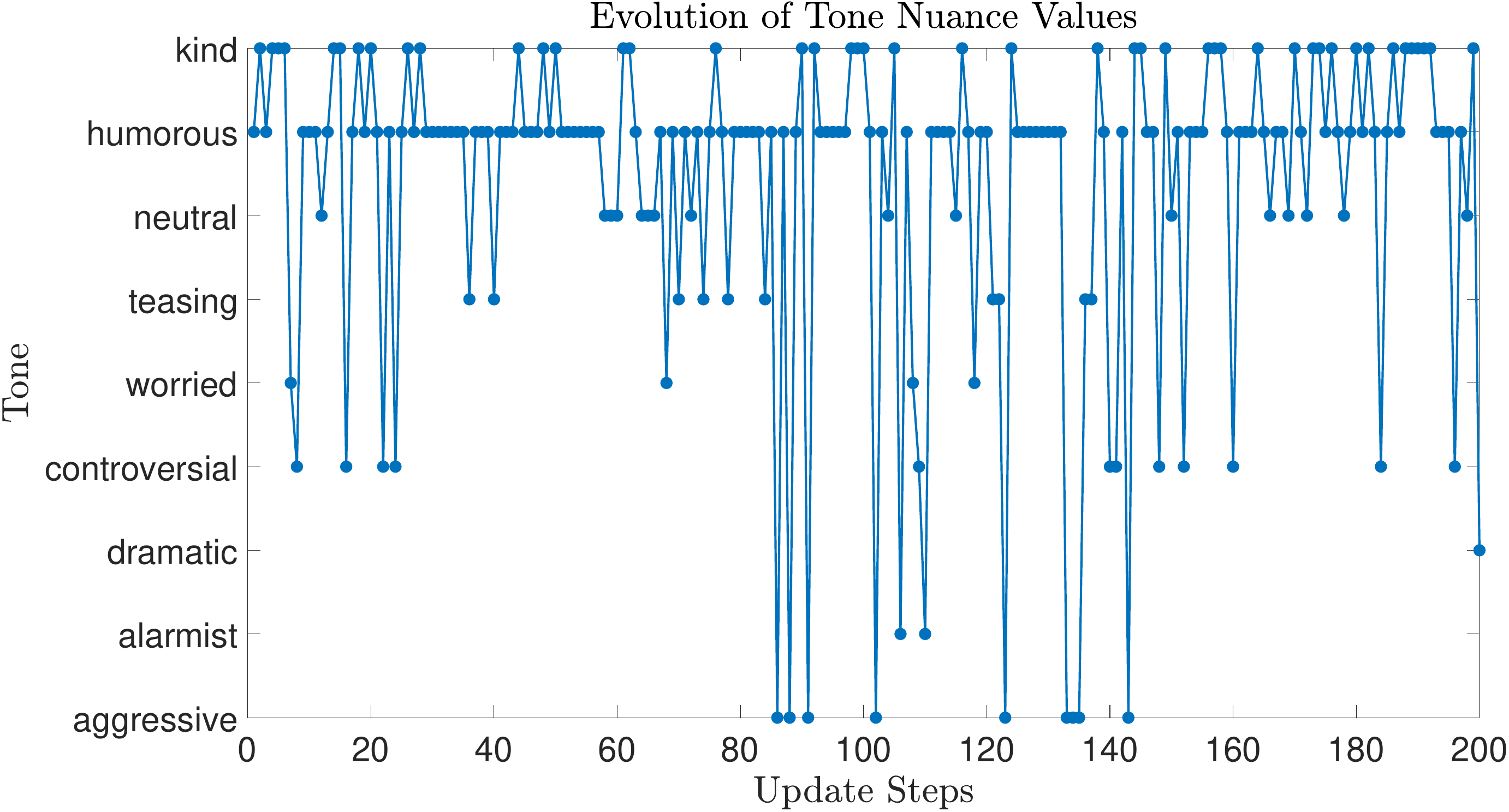}
	\caption{Evolution of the usage of tone nuance values across the first 100 turns using humorous sentences.}
	\vspace{-4mm}
  \label{fig:tone_nuance_values_humorous_controlled}
\end{figure}

Finally, Table \ref{tab:tone_nuances_controlled} presents the percentages of tone nuance usage throughout the entire experiment, considering the client sentences of the three specified tones. These reported values deviate from the steady-state distribution, reflecting the adjustments made in the nuance evolution whenever a humorous or aggressive tone is detected in the \textit{client sentence}. For instance, as depicted in Figure \ref{fig:tone_nuance_values_humorous_controlled}, the recognition of a humorous tone in the first 100 sentences correlates with an increase in the usage of the humorous tone nuance. Despite the varying percentages in the use of nuance values, the \textit{diversity cost} of tone nuance usage remains consistent between reply and continuation requests, requiring 12 tokens to specify the tone for the model's response generation. This accounts for approximately 1.5\% of both request prompts.

\begin{table}[tp]
	\centering
    \caption{Usage of tone nuance and steady-state distribution.}
	\begin{tabular}{c |p{0.8cm}p{1.5cm}p{1cm}p{0.7cm}}
		\toprule
		Tone nuance & Reply & Continuation & Overall & $\hat{\mathbf{e}}_{n}$ \\
		\midrule
		Humorous & 44.3\% & 21.0\% & 32.7\% & 0.092\\
		Kind & 24.7\% & 36.0\% & 30.3\% & 0.440\\
		Dramatic & 3.3\% & 2.0\% & 2.7\% & 0.060\\
		Controversial & 7.0\% & 10.3\% & 8.7\% & 0.145\\
		Aggressive & 7.0\% & 5.7\% & 6.3\% & 0.012\\
		Teasing & 2.7\% & 6.3\% & 4.5\% & 0.025\\
		Alarmist & 2.3\% & 4.0\% & 3.2\% & 0.060\\
		Worried & 2.3\% & 3.7\% & 3.0\% & 0.065\\
		Neutral & 6.3\% & 11.0\% & 8.7\% & 0.100\\
		\bottomrule
	\end{tabular}
 \vspace{-1mm}
	\label{tab:tone_nuances_controlled}
\end{table}

\subsubsection{GPT response times}
\label{sec:results_gpt_response_times_controlled}
Table \ref{tab:response_times_controlled} presents the average response times for the four requests performed to the language models, along with their corresponding standard deviations, minimum, and maximum values, all measured in seconds.

\begin{table}[tp]
	\centering
	\caption{Response times of the models.}
	\begin{tabular}{c |ccc}
		\toprule
		Request & Avg. resp. time ($\sigma$) [s] & Min [s] & Max [s] \\
		\midrule
		Sentiment request & 1.49 (0.89) & 0.58 & 7.06 \\
		Topic request & 1.50 (0.90) & 0.57 & 4.64\\
		Reply request & 3.17 (1.59) & 1.24 & 16.15 \\
		Continuation request & 2.39 (1.59) &  0.81 & 12.02 \\
		\bottomrule
	\end{tabular}
 \vspace{-1mm}
	\label{tab:response_times_controlled}
\end{table}

Note that the response times for both the \textit{sentiment request} and the \textit{topic request} exhibit lower values compared to the \textit{reply request} and \textit{continuation request}, aligning with the relative complexity of these requests.

\subsubsection{CAIR server response times}
When a client initiates a request, the Dialogue Manager concurrently sends three requests to language models, as depicted in Figure \ref{fig:sequence-diagram-llm}: \textit{sentiment}, \textit{topic}, and \textit{reply requests}. The server's response time is primarily influenced by the model's request that takes the longest, typically the \textit{reply request} (as indicated in Table \ref{tab:response_times_controlled}). This is evident from the average response time for the first request, which is 3.32$\pm$1.37~s, comprising the response time of the \textit{reply request} along with the time needed for network connectivity. Subsequently, upon receiving the first response, the client reproduces it while initiating a second request. The time for the second request is mainly influenced by the response time of the \textit{continuation request} to GPT, in addition to the network latency, averaging 2.49$\pm$1.35~s.

\subsubsection{CAIR client response time}
The client response at each turn is composed of three parts: a \textit{filler sentence}, the first part of the \textit{dialogue sentence} and the second part of the \textit{dialogue sentence}. The \textit{filler sentence} is used to signal to the user that the system has acknowledged that the user has finished talking. 
During the experiment, this happened after two seconds, which is the silence threshold chosen to detect the end of the user sentence. The other role of the \textit{filler sentence} is to fill the silence due to the time required to obtain the result of the audio acquisition plus the time required to obtain the response to the first request. Analysis of the experiment logs revealed that the audio acquisition result becomes available, on average, after 0.01$\pm$0.02~s once the system detects the end of the user's speech. After this acknowledgment, the client begins speaking the \textit{filler sentence} while initiating the first request to the CAIR server. As soon as the client receives the first response, it starts speaking the \textit{reply sentence}, while performing the second request. Finally, when the client receives the reply to the second request it starts reproducing the \textit{continuation sentence}.

\begin{figure}[pt]
	\centering
	\includegraphics[width=\linewidth]{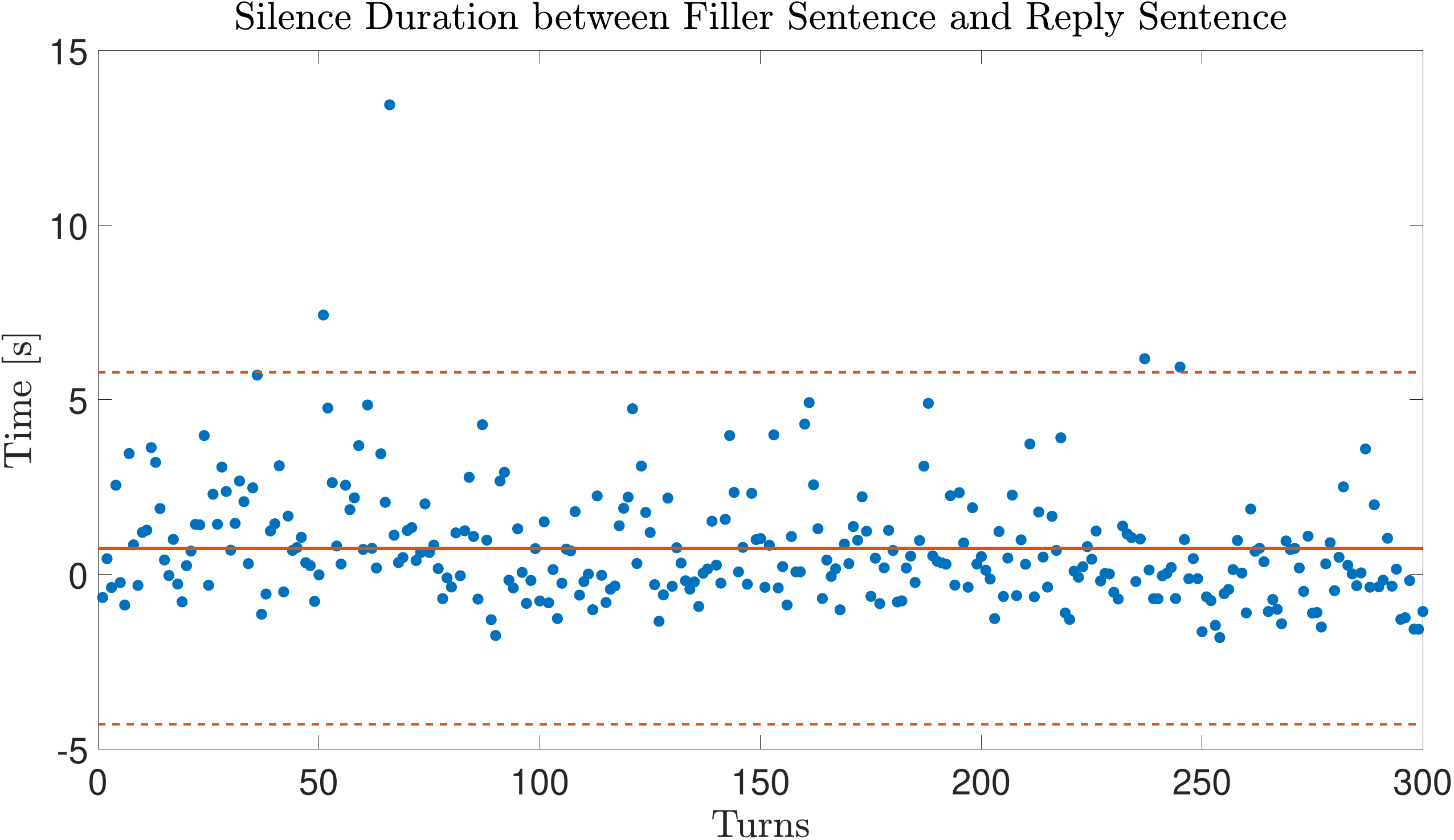}
	\caption{Time interval between the conclusion of the client's utterance of the \textit{filler sentence} and the commencement of the \textit{reply sentence}. The red line represents the average, while the dotted red line indicates $3\sigma$.}
	\label{fig:difference_first_response_time_filler_sentence_controlled}
    \vspace{-4mm}
\end{figure}

A crucial aspect is to measure the silence between the three sentences uttered by the robot. Let us define $t_1$ as the moment when the client finishes reproducing the \textit{filler sentence}, and $t_2$ as the moment when it begins uttering the \textit{reply sentence}, i.e., as soon as it receives the response to the first request. The time interval $t_2$-$t_1$ has been computed for all conversation turns and is depicted in Figure \ref{fig:difference_first_response_time_filler_sentence_controlled}, averaging 0.7 seconds. Similarly, the time interval between when the client finishes uttering the \textit{reply sentence} ($t_3$) and the moment it begins uttering the continuation sentence ($t_4$) has been calculated as $t_4$-$t_3$. The average was close to zero, as the time taken for the robot to reproduce the \textit{reply sentence} always exceeds the time taken to obtain the response to the second request made to the CAIR server.

\subsection{Diversity-aware robotics in the wild}
The 
system described in this work has been employed in two real-world case studies. Each case study presents a unique environment and set of challenges, offering valuable insights into 
its adaptability and performance in dynamic, public settings. For the sake of brevity, the detailed results of these case studies are not included in this work. 

\subsubsection{Case study 1}
We presented our system at the “Maker Faire'' in Rome and the “Festival della Scienza'' in Genoa, annual events that showcase science innovation and creativity. Our goal was to introduce and demonstrate our solution through the interactive capabilities of both Pepper and NAO robots. During both exhibitions, the system utilized the GPT-4 model to generate dialogue responses and relied on the GPT-3.5 Turbo model to determine the sentiment and topic of user sentences. Across seven days of exhibitions, visitors actively engaged with the robot, resulting in 1800 conversation turns exploring 244 different conversation topics. 

\subsubsection{Case study 2}
The second case study involves an ex-hospitalized paraplegic woman to assess how the system behaves when left alone in a home environment without the supervision of a researcher. For this purpose, the NAO robot was brought to the woman's house and remained there for six days. The first noteworthy result is that no technical problems occurred, and she was able to engage in conversation with the robot throughout the entire duration.
Throughout the six days, the woman engaged in 1003 dialogue turns, exploring 166 different conversation topics. Notably, the interactions persisted beyond the initial days, occurring consistently each day for substantial durations ranging from one to three hours. This indicates a sustained engagement, suggesting that even after the novelty wore off, she found the conversations with the robot stimulating. Furthermore, the feedback at the end of the testing period was highly positive, with the woman expressing that she developed a friendship with the robot. 

\begin{figure}[pt]
	\centering
	\includegraphics[width=0.8\linewidth]{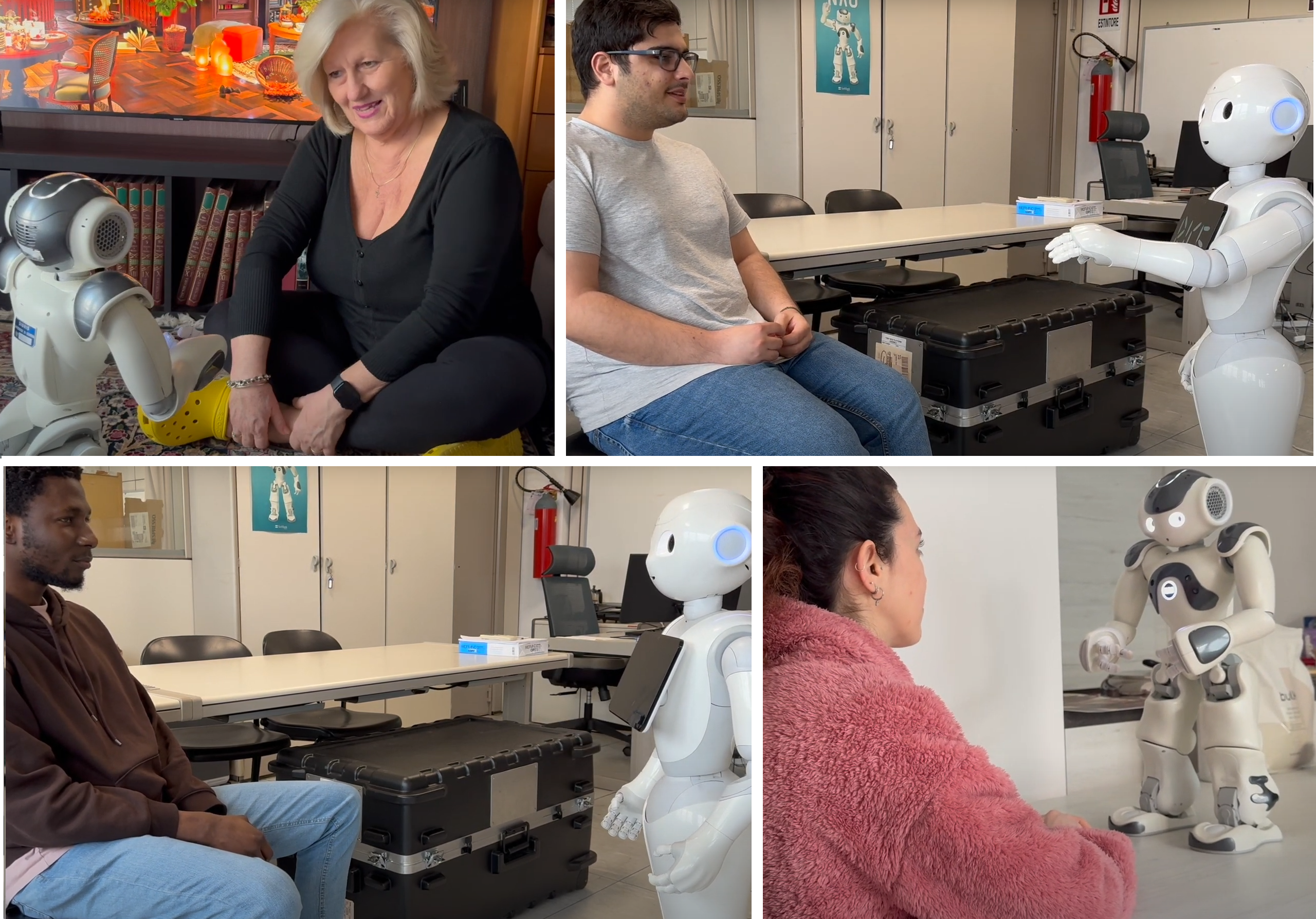}
	\caption{Diversity-aware interaction at home and in the RICE lab.}
	\label{fig:div}
 \vspace{-6mm}
\end{figure}

A video featuring diversity-aware interaction with various robots connected to the CAIR cloud in different environments and languages is available on YouTube\footnote{\url{https://youtu.be/FtaFB0iPl6w}} (Figure \ref{fig:div}).

\section{Conclusions}
\label{sec:conclusions}
This work presented the implementation of a diversity-aware conversational system integrating LLMs to enhance the ontology-driven dialogue capabilities. The system achieves this by constructing prompts using diversity-related information. This approach enables CAIR to guide conversations in alignment with the predefined structure of the ontology, maintaining control over the dialogue. The LLM is tasked with understanding the topic and the tone of the user sentence, evaluating preferences regarding conversation topics, and generating replies. These replies are crafted on specific topics, either identified by the LLM or selected by the system based on the structure of the knowledge base. This hybrid approach exploits the text generation capabilities of LLMs while relying on the ontology structure.

Performance evaluations, with a particular focus on the usage of diversity information and its associated cost in terms of tokens and response times, have been presented. Furthermore, preliminary real-world experiments have showcased the system's ability to engage in conversations across diverse settings, including crowded and noisy environments. Notably, the system has demonstrated consistent performance over an extended period in a home environment, operating without the need for technical assistance from developers.

\section{Acknowledgement}
\label{sec:acknowledgement}
This work has been supported by Ministero dell’Università e della Ricerca (Italian Ministry of University and Research), PNC - Piano Nazionale Complementare, FIT4MEDROB (PNC0000007) “Fit4MedRob - Fit for Medical Robotics'' Project - Mission 2

\addtolength{\textheight}{-12cm}   




\bibliographystyle{ieeetr}
\bibliography{bibliography}

\begin{thebibliography}{10}

\bibitem{martin2020}
A.~Martín, J.~Pulido, J.~González, A.~García-Olaya, and C.~Suárez, ``A framework for user adaptation and profiling for social robotics in rehabilitation,'' {\em Sensors}, vol.~20, no.~17, p.~4792, 2020.

\bibitem{parisi2019}
G.~Parisi, R.~Kemker, J.~Part, C.~Kanan, and S.~Wermter, ``Continual lifelong learning with neural networks: A review,'' {\em Neural Networks}, vol.~113, pp.~54--71, 2019.

\bibitem{buolamwini2018}
J.~Buolamwini and T.~Gebru, ``Gender shades: Intersectional accuracy disparities in commercial gender classification,'' {\em Conference on Fairness, Accountability, and Transparency}, 2018.

\bibitem{bolukbasi2016}
K.-W. Chang, J.~Zou, S.~V., and A.~Kalai, ``Man is to computer programmer as woman is to homemaker? debiasing word embeddings,'' {\em NIPS}, 2016.

\bibitem{mehrabi2021}
N.~Mehrabi, F.~Morstatter, N.~Saxena, K.~Lerman, and A.~Galstyan, ``A survey on bias and fairness in machine learning,'' {\em ACM Comput. Surv.}, vol.~54, jul 2021.

\bibitem{recchiuto2022}
C.~Recchiuto and A.~Sgorbissa, ``Diversity-aware social robots meet people: beyond context-aware embodied ai,'' in {\em Anthropology, AI and the Future of Human Society}, 2022.

\bibitem{vemprala2023chatgpt}
S.~Vemprala, R.~Bonatti, A.~Bucker, and A.~Kapoor, ``Chatgpt for robotics: Design principles and model abilities,'' {\em Microsoft Auton. Syst. Robot. Res}, vol.~2, p.~20, 2023.

\bibitem{takato2023}
T.~Yamazaki, K.~Yoshikawa, T.~Kawamoto, T.~Mizumoto, M.~Ohagi, and T.~Sato, ``Building a hospitable and reliable dialogue system for android robots: a scenario-based approach with large language models,'' {\em Adv. Rob.}, vol.~0, no.~0, pp.~1--18, 2023.

\bibitem{Billing2023}
E.~Billing, J.~Ros{\'e}n, and M.~Lamb, ``Language models for human-robot interaction,'' in {\em ACM/IEEE Proc. HRI'23}, (Stockholm, Sweden), pp.~905--906, March 2023.

\bibitem{Sonderegger2022}
S.~Sonderegger, ``How generative language models can enhance interactive learning with social robots.,'' {\em International Association for Development of the Information Society}, 2022.

\bibitem{zamfirescu2023johnny}
J.~D. Zamfirescu-Pereira, R.~Wong, B.~Hartmann, and Q.~Yang, ``Why {J}ohnny can’t prompt: how non-{AI} experts try (and fail) to design {LLM} prompts,'' in {\em Proc. of the CHI Conference}, pp.~1--21, 2023.

\bibitem{gao2023}
A.~Gao, ``Prompt engineering for large language models,'' {\em SSRN}, 2023.

\bibitem{liu2023prompt}
P.~Liu, W.~Yuan, J.~Fu, Z.~Jiang, H.~Hayashi, and G.~Neubig, ``Pre-train, prompt, and predict: a systematic survey of prompting methods in natural language processing,'' {\em ACM Comput. Surv.}, vol.~55, jan 2023.

\bibitem{grassi2023iros}
L.~Grassi, C.~Recchiuto, and A.~Sgorbissa, ``Robot-induced group conversation dynamics: A model to balance participation and unify communities,'' in {\em Proc. IEEE/RSJ IROS 2023}, (Detroit, USA), 2023.

\bibitem{recchiuto2020b}
C.~Recchiuto, L.~Gava, L.~Grassi, A.~Grillo, M.~Lagomarsino, D.~Lanza, Z.~Liu, C.~Papadopoulos, I.~Papadopoulos, A.~Scalmato, {\em et~al.}, ``Cloud services for culture aware conversation: Socially assistive robots and virtual assistants,'' in {\em Proc. UR'20}, pp.~270--277, 2020.

\bibitem{recchiuto2020a}
C.~T. Recchiuto and A.~Sgorbissa, ``A feasibility study of culture-aware cloud services for conversational robots,'' {\em IEEE Robot. Autom. Lett.}, vol.~5, no.~4, pp.~6559--6566, 2020.

\bibitem{grassi2023jist}
L.~Grassi, C.~T. Recchiuto, and A.~Sgorbissa, ``Sustainable cloud services for verbal interaction with embodied agents,'' {\em Intel Serv Robotics}, p.~599–618, 2023.

\bibitem{grassi2024ICRA}
L.~Grassi, Z.~Hong, C.~T. Recchiuto, and A.~Sgorbissa, ``Grounding conversational robots on vision through dense captioning and large language models,'' in {\em Proc. IEEE ICRA 2024}, 2024.

\end{thebibliography}
\end{document}